%% file: dpVAE-CVPR2020.tex
\documentclass[10pt,twocolumn]{article}

\input{preamble.tex}

\input{notations.tex}
\usepackage{enumitem}
\usepackage{multirow}
\usepackage{hhline}
\usepackage[toc,page]{appendix}

\begin{document}


 \title{\ipVAEs: Fixing Sample Generation for Regularized VAEs \thanks{This work was supported by the National Institutes of Health under the grant number NIAMS-R01AR076120. The content is solely the responsibility of the authors and does not necessarily represent the official views of the National Institutes of Health}}
\author{Riddhish Bhalodia, Iain Lee, Shireen Elhabian\\
Scientific Computing and Imaging Institute, \\School of Computing, University of Utah, Salt Lake City, UT, USA\\
	{\tt\small \{riddhishb, iclee, shireen\}@sci.utah.edu}}

\maketitle

\input{abstract_v7.tex}
\input{introduction_v5.tex}

\input{related_work_v2.tex}

\input{background_v5.tex}

\input{methods_v4.tex}

\input{results_v2.tex}
\input{conclusion.tex}

\appendix

\bibliographystyle{plain}
\bibliography{dpVAE-CVPR2020}

\section*{Appendix}
\input{appendix.tex}

\end{document}

%% file: preamble.tex
\usepackage{cvpr}
\usepackage{times}
\usepackage{epsfig}
\usepackage{graphicx}
\usepackage{amsmath}
\usepackage{amssymb}
\usepackage{dirtytalk}
\usepackage{xcolor}
\usepackage{cite}

\cvprfinalcopy 

\def\cvprPaperID{4521} 

\ifcvprfinal\fi

\allowdisplaybreaks
\expandafter\def\expandafter\normalsize\expandafter{%
	\normalsize
	\setlength\abovedisplayskip{2pt}
	\setlength\belowdisplayskip{2pt}
	\setlength\abovedisplayshortskip{2pt}
	\setlength\belowdisplayshortskip{2pt}
}

%% file: notations.tex
\newcommand{\stdNrmZ}{\mathcal{N}(\mathbf{z}; \mathbf{0}, \mathbb{I})}
\newcommand{\stdNrmZo}{\mathcal{N}(\mathbf{z}_0; \mathbf{0}, \mathbb{I})}
\newcommand{\NrmZ}[2]{\mathcal{N}(\mathbf{z}; #1, #2)}

\newcommand{\muZX}{\boldsymbol{\mu}_\mathbf{z}(\mathbf{x})}
\newcommand{\SigmaZX}{\boldsymbol{\Sigma}_\mathbf{z}(\mathbf{x})}
\newcommand{\sigmaZXdiag}{\operatorname{diag}(\boldsymbol{\sigma}_\mathbf{z}(\mathbf{x}))}
\newcommand{\sigmaZX}{\boldsymbol{\sigma}_\mathbf{z}(\mathbf{x})}

\renewcommand{\figurename}[0]{Figure}
\newcommand{\rfigure}[1]{\figurename~\ref{#1}}
\newcommand{\requation}[1]{(\ref{#1})}

\newcommand{\aka}[0]{a.k.a.~}

\newcommand{\sota}[0]{state-of-the-art~}
\newcommand{\Sota}[0]{State-of-the-art~}

\newcommand{\ipVAE}[0]{\textit{dp}VAE}
\newcommand{\ipVAEs}[0]{\textit{dp}VAEs}

\newcommand{\x}[0]{\mathbf{x}}
\newcommand{\z}[0]{\mathbf{z}}
\newcommand{\zo}[0]{\mathbf{z}_0}
\newcommand{\zk}[1]{\mathbf{z}_{#1}}
\newcommand{\zkl}[2]{z_{#1}^{#2}}
\newcommand{\zl}[1]{z^{#1}}

\newcommand{\bk}[1]{\mathbf{b}_{#1}}
\newcommand{\bkl}[2]{b_{#1}^{#2}}

\newcommand{\ptheta}[0]{p_\theta}

\newcommand{\qphi}[0]{q_\phi}

\newcommand{\RL}[0]{\mathbb{R}^L}
\newcommand{\RLp}[0]{\mathbb{R}^L_+}
\newcommand{\RD}[0]{\mathbb{R}^D}

\newcommand{\mcL}[0]{\mathcal{L}}
\newcommand{\mcZ}[0]{\mathcal{Z}}
\newcommand{\mcZo}[0]{\mathcal{Z}_0}

\newcommand{\gdefine}[0]{g:\mathcal{Z} \rightarrow \mathcal{Z}_0}
\newcommand{\g}[1]{g(#1)}
\newcommand{\geta}[1]{g_\eta(#1)}
\newcommand{\getak}[1]{g_\eta^{(#1)}}
\newcommand{\getainv}[1]{g_\eta^{#1}}
\newcommand{\getakz}[2]{g_\eta^{(#1)}(#2)}
\newcommand{\getadef}[0]{g_\eta}

\newcommand{\ginvdefine}[0]{g^{-1}:\mathcal{Z}_0 \rightarrow \mathcal{Z}}
\newcommand{\ginv}[1]{g^{-1}(#1)}
\newcommand{\ginveta}[1]{g_\eta^{-1}(#1)}

\newcommand{\KL}[2]{\operatorname{KL}\left[ #1 \| #2 \right]}
\newcommand{\E}[2]{\mathbb{E}_{#1}\left[#2\right]}
\newcommand{\En}[2]{\mathbb{E}_{#1}[#2]}

\renewcommand{\and}[0]{\operatorname{and}}

%% file: abstract_v7.tex
\begin{abstract}
Unsupervised representation learning via generative modeling is a staple to many computer vision  applications in the absence of labeled data.
%
Variational Autoencoders (VAEs) are powerful generative models that learn representations useful for data generation.
However, due to inherent challenges in the training objective, VAEs fail to learn useful representations amenable for downstream tasks. 
Regularization-based methods that attempt to improve the representation learning aspect of VAEs come at a price: poor sample generation. 
In this paper, we explore this representation-generation trade-off 
for regularized VAEs and introduce a new family of 
priors, namely \textit{decoupled priors}, or \ipVAEs, that decouple the representation space from the generation space. 
This decoupling enables the 
use of VAE regularizers on the representation space without impacting the distribution used for sample generation, and thereby reaping the representation learning benefits of the regularizations without sacrificing the sample generation.
%
\ipVAE~ leverages invertible networks to learn a bijective mapping from an arbitrarily complex representation distribution to a simple, tractable, generative distribution.  Decoupled priors can be adapted to the \sota VAE regularizers without additional hyperparameter tuning. We showcase the use of \ipVAEs~ with different regularizers. 
Experiments on MNIST, SVHN, and CelebA demonstrate, quantitatively and qualitatively, that \ipVAE~ fixes sample generation for regularized VAEs.
\end{abstract}

%% file: introduction_v5.tex
\section{Introduction}\label{sec:introduction}
\begin{figure}
    \centering
    \includegraphics[width=\linewidth]{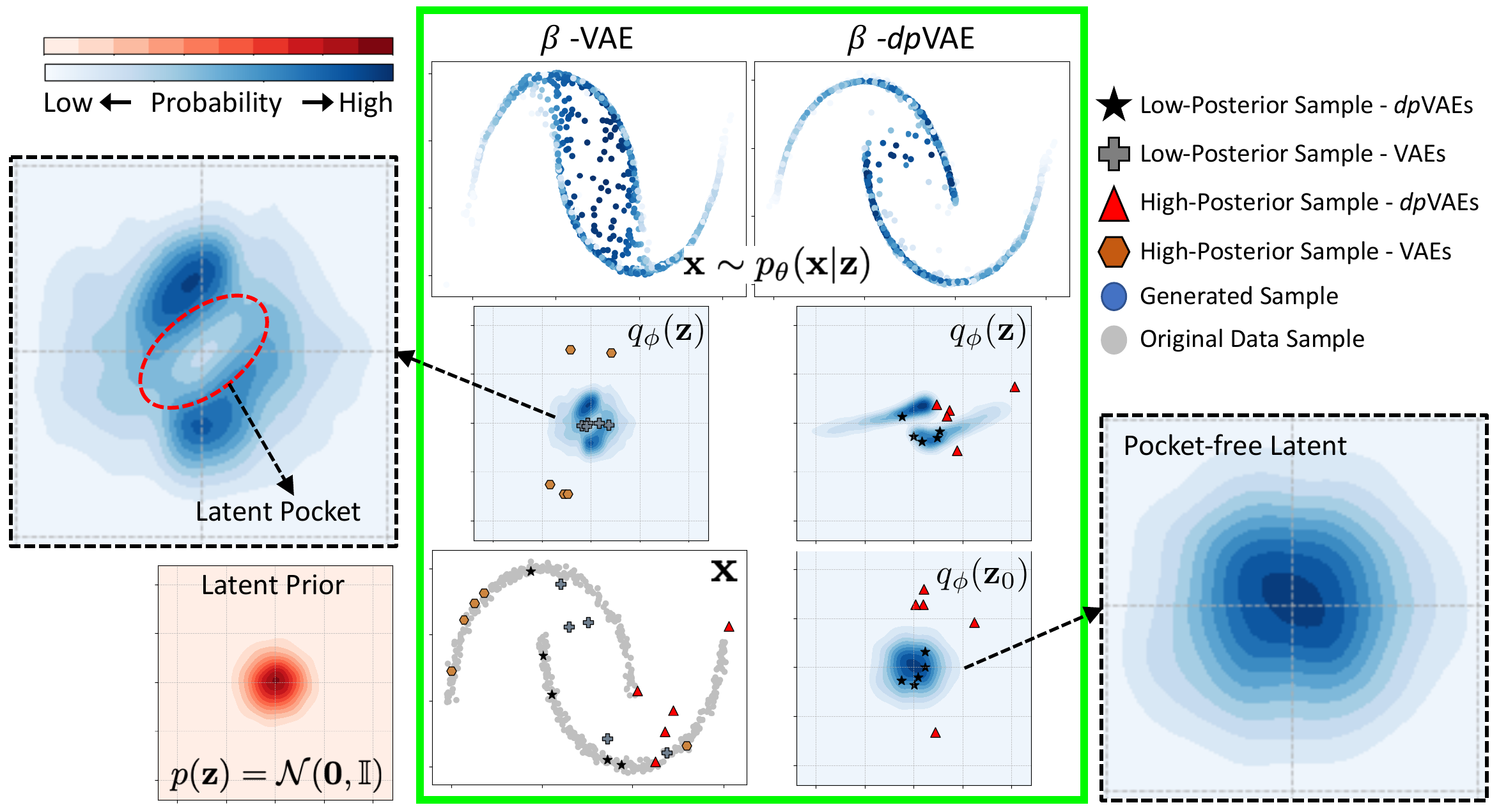}
        \caption{ \textbf{\ipVAE~ fixes sample generation for a regularized VAE.}
        %
        The \textcolor{green}{green box} shows $\beta$-VAE \cite{higgins2017beta} (left column) and $\beta$-VAE with the proposed \emph{decoupled prior} (right column), each trained on the two moons dataset.
        $\beta$-VAE: Top to bottom shows the generated samples (colors reflect probability of generation), the aggregate posterior $\qphi(\z)$ and the training samples. The low-posterior samples lie in the latent pockets of $\qphi(\z)$ (shown in enlarged section on the left) and correspond to off-manifold samples in the data space, and high-posterior samples correspond to latent leaks. The $\beta$-\ipVAE~ decouples the representation $\z$ and generation $\zo$ spaces. The generation space is pocket-free and very close to standard normal, resulting in generating samples on the data manifold. Furthermore, the representation learning is well established in the representation space (see section \ref{sec:submanifold} for more discussion).
        }
    \label{fig:teaser}
\end{figure}
%
\say{Is it possible to learn a \textit{powerful generative model} that matches the true data distribution with \textit{useful  data representations} amenable to downstream tasks in an unsupervised way?} 
---This question is the driving force behind most unsupervised representation learning via \sota generative modeling methods (\eg \cite{zhao2019infovae,higgins2017beta,kim2018disentangling,chen2016infogan}), with applications in artificial creativity \cite{nguyen2017plug,mathieu2016disentangling}, reinforcement learning \cite{higgins2017darla}, few-shot learning \cite{rezende2016one}, and semi-supervised learning \cite{kingma2014semi}. 
A common theme behind such works is learning the data generation process using \textit{latent variable models} \cite{bengio2013representation,alemi2018fixing} that seek to learn representations useful for data generation; an approach known as \textit{analysis-by-synthesis} \cite{yuille2006vision,nair2008analysis}.

%
The Variational Autoencoder (VAE)  \cite{kingma2014auto,rezende2014stochastic} marries latent variable models and deep learning by having independent, network-parameterized \textit{generative} and \textit{inference} models that are trained jointly to maximize the marginal log-likelihood of the training data.  
VAE introduces a variational posterior distribution that approximates the true posterior to derive a tractable lower bound on the marginal log-likelihood, \aka the evidence lower bound (ELBO). 
%
The ELBO is then maximized using stochastic gradient descent by virtue of the reparameterization trick \cite{kingma2014auto,rezende2014stochastic}.
Among the many successes of VAEs in representation learning tasks, VAE-based methods have demonstrated state-of-the-art performance for semi-supervised image and text classification tasks \cite{maaloe2016auxiliary,kingma2014semi,sonderby2016train,rezende2016one,pu2016variational,xu2017variational}. 

%
Representation learning via VAEs is ill-posed due to the disconnect between the ELBO and the downstream task \cite{tschannen2018recent}. Specifically, optimizing the marginal log-likelihood is not always sufficient for good representation learning due to the inherent challenges rooted in the ELBO that result in the tendency to \textit{ignore latent variables} and \textit{not encode information about the data in the latent space} \cite{tschannen2018recent,alemi2018fixing,zhao2019infovae,chen2017variational,hoffman2016elbo}.  
%
To improve the representations learned by VAEs, a slew of regularizations have been proposed. Many of these regularizers act on the VAE latent space to promote specific characteristics in the learned representations, such as disentanglement  \cite{zhao2019infovae,higgins2017beta,kim2018disentangling,mathieu2016disentangling,chen2018isolating,kumar2018variational} and informative latent codes \cite{makhzani2017pixelgan,alemi2016deep}.
However, better representation learning usually sacrifices sample generation, which is manifested by a distribution mismatch between the marginal (\aka aggregate) latent posterior 
and the latent prior. 
This mismatch results in \textit{latent pockets and leaks}; a \textit{submanifold} structure in the latent space (a phenomena demonstrated in \rfigure{fig:teaser} and explored in more detail in section \ref{sec:submanifold}). Latent pockets contain samples that are highly supported under the prior but not covered by the aggregate posterior (\ie \textit{low-posterior samples} \cite{rosca2018distribution}), while latent leaks contain samples supported under the aggregate posterior but less likely to be generated under the prior (\ie \textit{high-posterior samples}). 
%
This behavior has been reported for 
VAE \cite{rosca2018distribution} but it is substantiated by augmenting the ELBO with regularizers (see \rfigure{fig:teaser}).

To address this representation-generation trade-off for regularized VAEs, we introduce the idea of decoupling the latent space for representation (\textit{representation space}) from the space that drives sample generation (\textit{generation space}); presenting 
a general framework for VAE regularization. 
To this end, we propose a new family of latent priors for VAEs --- \textit{decoupled priors} or \ipVAEs~ --- that leverages the merits of invertible deep networks. 
%
%
In particular, \ipVAE~ transforms a tractable, simple base prior distribution 
in the generation space to a  more expressive prior in the representation space that reflects the submanifold structure dictated by the regularizer. This is done using an invertible mapping that is jointly trained with the VAE's inference and generative models. \Sota VAE regularizers can thus be directly plugged in to promote specific characteristics in the representation space without impacting the distribution used for sample generation.
%
We showcase, quantitatively and qualitatively, that \ipVAE~ with different \sota regularizers improve sample generation, without sacrificing their representation learning benefits.
%

It is worth emphasizing that, being likelihood-based models, VAEs are trained to put probability mass on all training samples, forcing the model to \textit{over-generalize} \cite{shmelkov2019coverage}, and generating blurry samples (\ie off data manifold).
%
This is in contrast to generative adversarial networks (GANs) \cite{goodfellow2014generative} that generate outstanding image quality but lack the full data support
\cite{arjovsky2017wasserstein}. \ipVAE~ is not expected to resolve the over-generalization problem in VAEs, but to mitigate poor sample quality resulting from regularization.

\vspace{0.05in}
\noindent The contribution of this paper is fourfold:
%
\begin{itemize}
    \item Analyze the latent submanifold structure induced by VAE regularizers.
    \item Introduce a decoupled prior family for VAEs as a general regularization framework that improves sample generation without sacrificing representation learning. 
    
    \item Derive the \ipVAE~ ELBO of \sota regularized VAEs; $\beta$-\ipVAE, $\beta$-TC-\ipVAE, Factor-\ipVAE, and Info-\ipVAE.

    \item Demonstrate empirically on three benchmark datasets the improved generation performance and the preservation of representation characteristics promoted via regularizers without additional hyperparameter tuning.
\end{itemize}
%

%% file: related_work_v2.tex
\section{Related Work} \label{sec:related-work}

To improve sample quality, a family of approaches exist that combine the inference capability of VAEs and the outstanding sample quality of GANs  \cite{goodfellow2014generative}.
Leveraging the density ratio trick \cite{goodfellow2014generative,sugiyama2012density} that only requires samples, VAE-GAN hybrids in the latent (\eg \cite{mescheder2017adversarial,makhzani2015adversarial}), data (\eg \cite{rosca2018distribution,shmelkov2019coverage}), and joint (both latent and data \eg \cite{srivastava2017veegan}) spaces avoid restrictions to explicit posterior and/or likelihood distribution families, paving the way for marginals matching \cite{rosca2018distribution}.
However, such hybrids scale poorly with latent dimensions, lack accurate likelihood bound estimates, and do not provide better quality samples than GAN variants \cite{rosca2018distribution}.
%
Expressive posterior distributions can lead to better sample quality \cite{mescheder2017adversarial,kingma2016improved} and are essential to prevent latent variables from being ignored in case of powerful generative models \cite{chen2017variational}. But results in \cite{rosca2018distribution} suggest that the posterior distribution is not the main learning roadblock for VAEs.
%

%
More recently, the key role of the prior distribution family in VAE training has been investigated  \cite{hoffman2016elbo,rosca2018distribution}; poor latent representations are often attributed to restricting the latent prior to an overly simplistic distribution (\eg standard normal). 
This motivates several works to enrich VAEs with more expressive priors.
%
Bauer and Mnih addressed the distribution mismatch between the aggregate posterior and the latent prior by learning a sampling function, parameterized by a neural network, in the latent space  \cite{bauer2019resampled}. However, this resampled prior requires the estimation of the normalization constant and dictates an inefficient iterative sampling, where a truncated sampling could be used at the price of a less expressive prior due to smoothing. 
%
Tomczak and Welling proposed the variational mixture of posteriors prior (VampPrior), which is a parameterized mixture distribution in the latent space given by a fixed number of learnable pseudo (\ie virtual) data points \cite{tomczak2018vae}. VampPrior sampling is non-iterative and is therefore fast. However, density evaluation is expensive due to the requirement of a large number of pseudo points, typically in the order of hundreds, to match the aggregate posterior \cite{bauer2019resampled}.
%
A cheaper version is a mixture of Gaussian prior proposed in \cite{dilokthanakul2016deep}, which gives an inferior performance compared to VampPrior and is more challenging to optimize \cite{bauer2019resampled}.
%
Autoregressive priors (\eg \cite{gregor2015draw,gulrajani2017pixelvae}) come with fast density evaluation but a slow, sequential sampling process.
With differences between expressiveness and efficiency, none of these methods address the fundamental challenge of VAE training in concert with existing representation-driven regularization frameworks.

The proposed decoupled prior is inspired by flow-based generative models \cite{kingma2016improved,dinh2014nice,dinh2017density,rezende2015NF}, which have shown their efficacy in generating images (\eg GLOW \cite{kingma2018glow}). Such methods hinge on architectural designs that make the model invertible. These models are not used for representation learning though, since the data and latent spaces are required to have the same dimension.

%% file: background_v5.tex
\section{Background} \label{sec:background}



In this section, we briefly lay down the foundations and motivations essential for the proposed VAE formulation.

\subsection{Variational Autoencoders}
\label{sec:VAE}


%
%

VAE seeks to match the learned model distribution $\ptheta(\x)$ to the true data distribution $p(\x)$, where $\x \in \RD$ is the observed variable in the 
data space. The generative and inference models in VAEs are thus jointly trained to maximize a tractable lower bound $\mcL(\theta,\phi)$ on the marginal log-likelihood $\E{p(\x)}{\log \ptheta(\x)}$ of the training data, where $\z \in \RL$ is an unobserved latent variable in the latent space with a  prior distribution $p(\z)$, such as $p(\z) \sim \stdNrmZ$.
%
\begin{align}\label{eq:elbo-reconstruction-minus-priorKL}
    \mcL(\theta,\phi) = \E{p(\x)}{ \E{\qphi(\z|\x)}{\log \ptheta(\x|\z)} \right. \nonumber\\  \left. - \KL{\qphi(\z|\x)}{p(\z)}}
\end{align}
\noindent where $\theta$ denotes the generative model parameters, $\phi$ denotes the inference model parameters, and $\qphi(\z|\x) \sim \NrmZ{\muZX}{\SigmaZX}$ is the variational posterior distribution that approximates the true posterior $p(\z|\x)$, where $\muZX \in \RL$, $\SigmaZX = \sigmaZXdiag$, and $\sigmaZX \in \RLp$.



%
Since the ELBO seeks to match the marginal data distribution without penalizing the poor quality of latent representation, VAE can easily ignore latent variables if a sufficiently expressive generative model $\ptheta(\x | \z)$ is used (\eg PixelCNN \cite{van2016conditional}) and still maximize the ELBO \cite{alemi2018fixing,bowman2016generating,chen2017variational}, a property known as \textit{information preference} \cite{chen2017variational,zhao2019infovae}. 
%
Furthermore, VAE has the tendency to not encode information about the observed  data in the latent codes since maximizing the ELBO is inherently minimizing the mutual information between $\z \sim \qphi(\z|\x)$ and $\x$ \cite{hoffman2016elbo}.
Without further 
assumptions or 
inductive biases, these failure modes hinder learning useful representations for downstream tasks. 

\subsection{Invertible Deep Networks}
\label{sec:ACB}

The proposed decoupled prior family for VAEs leverages flow-based generative models that are formed by a sequence of \textit{invertible} blocks (\ie transformations), parameterized by deep networks.
%
%
Consider two random variables $\z \in \mcZ \subset \RL$ and $\zo \in \mcZo \subset \RL$. There exist a bijective mapping between $\mcZ$ and $\mcZo$ defined by a function $\gdefine$, where $\g{\z} = \zo$, and its inverse $\ginvdefine$ such that $\z = \ginv{\zo}$. 
Given the above condition, we can define the \textit{change of variable formula} for mapping probability distribution on $\z$ to $\zo$ as follows:
\begin{align}
    p(\z) = p(\zo)\bigg |\frac{\partial \zo}{\partial \z}\bigg | = p(g(\z))\bigg | \frac{\partial g(\z)}{\partial \z}\bigg |
    \label{eq:realNVP}
\end{align}

By maximizing the log-likelihood and parameterizing the invertible blocks with deep networks, flow-based methods learn to transform a simple, tractable base distribution (\eg standard normal) into a more expressive one.
To model distributions with arbitrary dimensions, the $g-$bijection needs to be defined such that the Jacobian determinant can be computed in a closed form. Dinh \etal \cite{dinh2017density} proposed the \textit{affine coupling layers} to build a flexible bijection function $g$ by stacking a sequence of $K$ simple bijection blocks $\zk{k-1} = \getakz{k}{\zk{k}}$ of the form,
\begin{eqnarray}
\getakz{k}{\zk{k}}&=& \bk{k} \odot \zk{k}  \nonumber \\ 
&+& (1 - \bk{k}) \odot \left[ \zk{k} \odot \exp \left(s_k(\bk{k} \odot \zk{k}) \right) \right. \nonumber \\ 
&& ~~~~~~~~~~~~~~~~~~~ \left. + t_k\left(\bk{k} \odot \zk{k}\right)\right] \label{eq:invblock}
\end{eqnarray}
\begin{align}
    \geta{\z} = \zo = \getak{1} \circ \dots \circ \getak{K-1} \circ \getakz{K}{\z} \label{eq:gz}
\end{align}
\noindent where $\z = \zk{K}$, $\odot$ is the Hadamard (\ie element-wise) product,
$\bk{k} \in \{0,1\}^L$ is a binary mask used for partitioning the $k-$th block input, and $\eta = \{s_1, ..., s_K, t_1, ..., t_K\}$ are the deep networks parameters of the scaling $s_k$ and translation $t_k$ functions of the $K$ blocks. 
%

%% file: methods_v4.tex

\begin{figure}
    \centering
    \includegraphics[width=\linewidth]{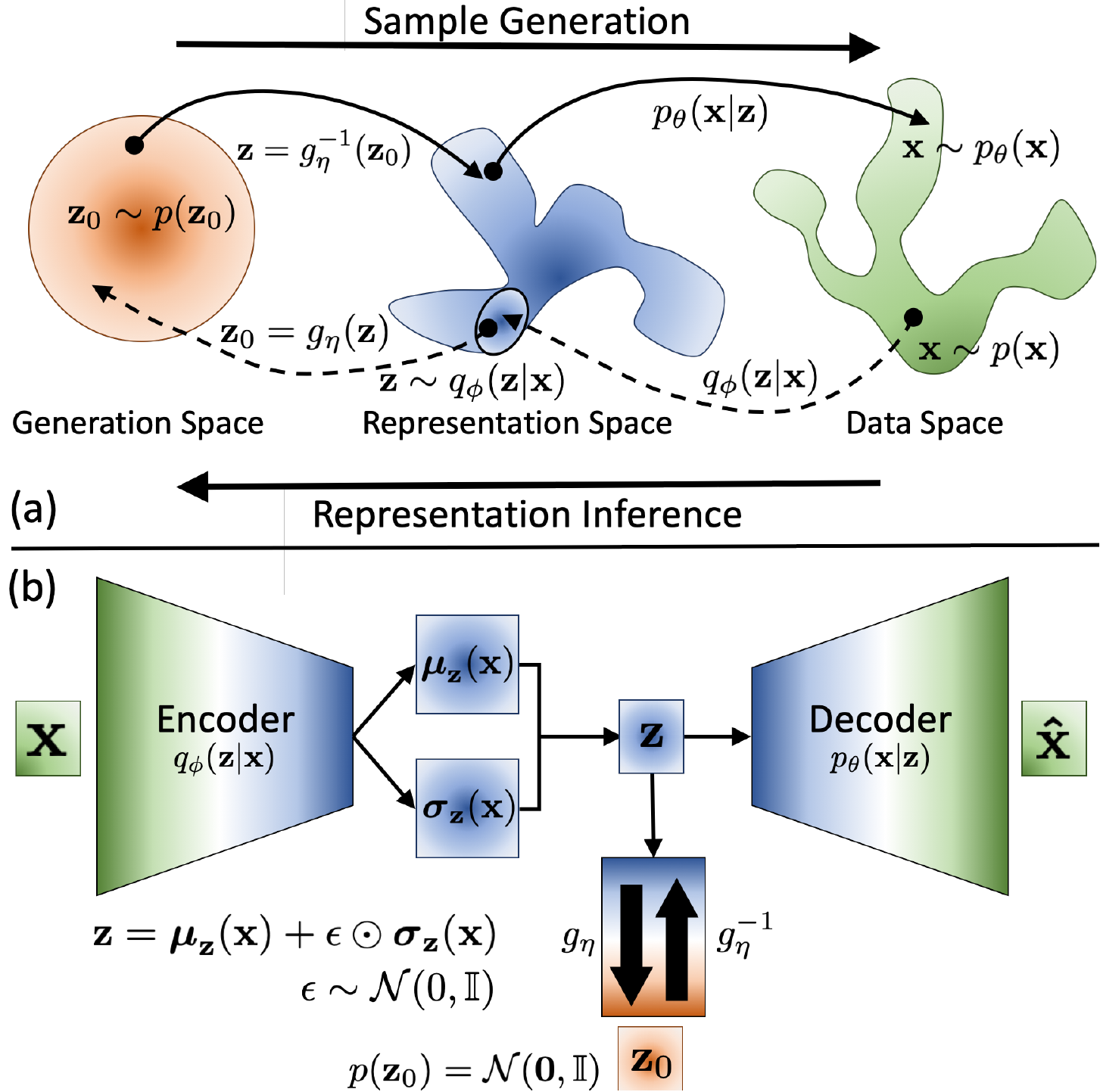}
    \caption{\textbf{\ipVAE:} (a) The latent space is decoupled into a generation space with a simple, tractable distribution (\eg standard normal) and a representation space whose distribution can be arbitrarily complex and is learned via a bijective mapping to the generation space. (b) Architecture of a VAE with the decoupled prior. The $g-$bijection is jointly trained with the VAE generative (\ie decoder) and inference (\ie encoder) models. 
    }
    \label{fig:IP-arch}
\end{figure}

\section{General Framework for VAE Regularization}\label{sec:methods}

In this section, we formally define and analyze how VAE regularizations affect the generative property of VAE. We also present the decoupled prior family for VAEs (see \rfigure{fig:IP-arch}) and analyze its utility to solve the submanifold problem of \sota regularization-based VAEs.

\begin{figure*}[!t]
    \centering
    \includegraphics[width =\textwidth]{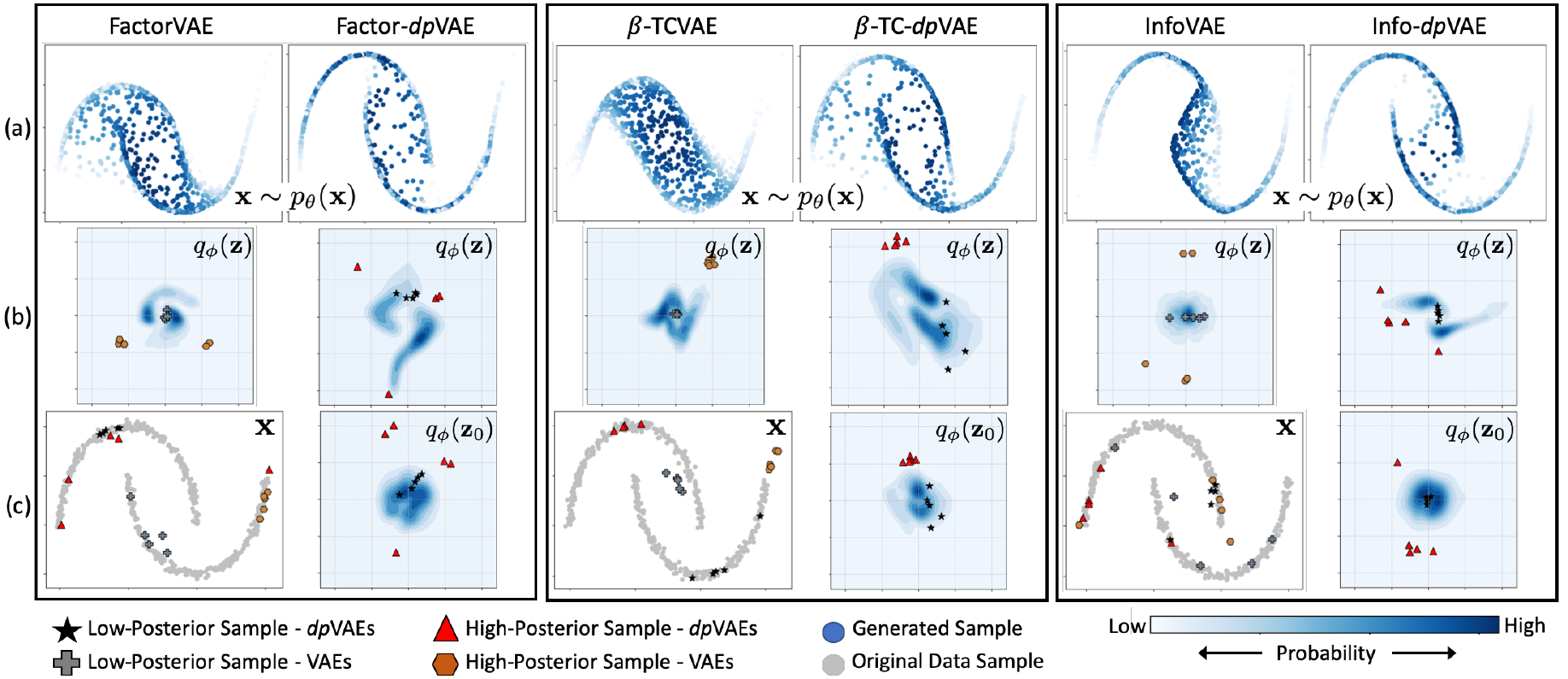}
    \caption{\textbf{Sample generation and latent spaces for regularized VAEs:} Each block is a VAE trained with a  different regularizer on the two moons dataset, with and without the decoupled prior. In each block, (a) showcases the sample generation quality, (b) shows the aggregate posterior $\qphi(\z)$ with top five low- and high-posterior samples marked, and (c) shows the generation space for the decoupled prior and the training samples in the data space with corresponding low- and high-posterior samples are marked.}
    \label{fig:moons_fails}
\end{figure*}

\subsection{VAE Regularizers: Latent Pockets and Leaks}
\label{sec:submanifold}

%
%
ELBO regularization is a conventional mechanism that enforces inductive biases (\eg disentanglement  \cite{zhao2019infovae,higgins2017beta,kim2018disentangling,mathieu2016disentangling,chen2018isolating,kumar2018variational} and informative latent codes \cite{makhzani2017pixelgan,alemi2016deep}) to improve the representation learning aspect of VAEs \cite{tschannen2018recent}. 
These methods have shown their efficacy in learning good representations but neglect the generative property.
Empirically, these regularizations improve the learned latent representation but inherently cause a mismatch between the aggregate posterior $\qphi(\z) = \E{p(\x)}{\qphi(\z | \x)}$ and the prior $p(\z)$. 
This mismatch leads to \emph{latent pockets and leaks}, or a \emph{submanifold} in the aggregate posterior that results in poor generative capabilities. Specifically, if a sample $\z \sim p(\z)$ (\ie likely to be generated under the prior) lies in a pocket, (\ie $\qphi(\z)$ is low), then its corresponding decoded sample $\x \sim \ptheta(\x|\z)$ will not lie on the data manifold. 
This problem, caused by VAE regularizations, we call the \emph{submanifold problem}.


To better understand this phenomena, we define two different types of samples in the VAE latent space that corresponds to two VAE failure modes.
    
\vspace{0.05in}
\noindent \textbf{Low-Posterior (LP) samples}: 
    These are samples that are highly likely to be generated under the prior (\ie $p(\z)$ is high) but are not covered by the aggregate posterior (\ie $\qphi(\z)$ is low). The low-posterior samples are typically generated from the \textit{latent pockets} dictated by the regularizer(s) used and are of poor quality since they lie off the data manifold.
    To generate low-posterior samples, we follow the logic of \cite{rosca2018distribution}, where we sample $\z \sim p(\z) = \stdNrmZ$, rank them according to their aggregate posterior support, \ie values of $\qphi(\z)$, and choose the samples with lowest aggregate posterior values. In the case of \ipVAEs, samples are generated from $\zo \sim p(\zo) =  \stdNrmZo$, which is a standard normal, and then transformed by $\z = \ginveta{\zo}$ before plugging it into the aggregate posterior.
    %
    
    \vspace{0.05in}
\noindent \textbf{High-Posterior (HP) samples}: These are samples supported under the aggregate posterior (\ie $\qphi(\z)$ is high) but are less likely to be generated under the prior (\ie $p(\z)$ is low). Specifically, these are samples in the latent space that can produce good generated samples but are unlikely to be sampled due to the low support of the prior, and thereby they are samples that are in the \textit{latent leaks}. 
    To generate high-posterior samples, we sample from $\z \sim \qphi(\z) = \E{p(\x)}{\qphi(\z | \x)}$, rank them according to their prior support, \ie values of $\stdNrmZ$, and choose the samples with lowest prior support values. In the case of \ipVAEs,  sampled $\z$ are first mapped to the $\zo-$space by $\zo = \geta{\z}$ before computing prior probabilities $\stdNrmZo$.

In summary, a VAE performs well in the generative sense if the latent space is free of pockets and leaks. 
A pocket-free latent space is manifested by low-posterior samples that lie on the data manifold when mapped to the data space via the decoder $\ptheta(\x | \z)$. 
In a leak-free latent space, high-posterior samples are supported by the aggregate posterior, yet with a tiny probability under the prior, and thereby these samples fall off the data manifold.
%
%
This submanifold problem is demonstrated using four \sota VAE regularizers (see \rfigure{fig:teaser} and \rfigure{fig:moons_fails}).
With $\beta$-VAE \cite{higgins2017beta}, FactorVAE \cite{kim2018disentangling} and $\beta$-TCVAE \cite{chen2018isolating}, we can clearly see that the low-posterior samples lie in the latent pockets formed in the aggregate posterior (see \rfigure{fig:moons_fails}b) and they lie outside the data manifold (see \rfigure{fig:moons_fails}c), causing the sample generation to be very noisy (see \rfigure{fig:moons_fails}a). 
In the case of InfoVAE, the low-posterior samples lie in regions with not much aggregate posterior support causing a slightly noisy sample generation (see \rfigure{fig:moons_fails}a).
More importantly, there are high-posterior samples that come from $\qphi(\z)$ but can very rarely be captured by a standard normal prior distribution. With the InfoVAE, for instance, the model fails to generate samples that lie on the tail-end of the top moon.

Although VAE regularizers improve latent representations, they sacrifice sample generation through the introduction of latent pockets and leaks.
To fix sample generation, 
we propose a decoupling of the representation and generation spaces
(see \rfigure{fig:IP-arch}a for illustration). This is demonstrated for $\beta$-VAE with and without decoupled prior in \rfigure{fig:teaser}, where the decoupled generation space $p(\zo) \sim \stdNrmZo$ is used for generation and all the low-posterior samples lie on the data manifold. We formulate this prior in detail in following section.
 
\subsection{\ipVAE: Decoupled Prior for VAE}
\label{sec:invertibleprior}

\emph{Decoupled prior} family, as the name suggests, decouples the latent space that performs the representation and the space that drives sample generation.  For this decoupling to be meaningful, the representation and generation spaces should be related by a functional mapping.
%
The decoupled prior effectively learns the latent space distribution $p(\z)$ by simultaneously learning the functional mapping $g_\eta$ together with the generative and inference models during optimization.


Specifically, the latent variables $\z \in \mcZ \subset \RL$ and $\zo \in \mcZo \subset \RL$ are the random variables of the \textit{representation} and \textit{generation} spaces, respectively, where $p(\zo) \sim \stdNrmZo$. 
The bijective mapping between the representation space $\mcZ$ and the generation space $\mcZo$ is defined by an invertible function $\geta{\z} = \zo$, parameterized by the network parameters $\eta$.
VAE regularizers still act on the posteriors in the representation space, \ie $\qphi(\z|\x)$ and/or $\qphi(\z)$, without affecting the latent distribution of the generation space $p(\zo)$. Sample generation starts by sampling $\zo \sim p(\zo)$ 
, passing through the inverse mapping to obtain $\z = \ginveta{\zo}$, which is then decoded by the generative model $\ptheta(\x | \z)$ (see \rfigure{fig:IP-arch}a). 
These decoupled spaces can allow any modifications in the representation space dictated by the regularizer to infuse its submanifold structure in that space (see \rfigure{fig:moons_fails}b) without significantly impacting the generation space (see \rfigure{fig:moons_fails}c), and thereby improving sample generation for regularized VAEs (see \rfigure{fig:moons_fails}a). Moreover, the decoupled prior $p(\z)$ is an expressive prior that is learned jointly with the VAE, and thereby it can match an arbitrarily complex aggregate posterior $\qphi(\z)$, thanks to the flexibility of deep networks to model complex mappings. Additionally, due to the bijective mapping $\getadef$, we have a one-to-one correspondence between samples in $p(\zo) \sim \stdNrmZo$ and those in $p(\z)$.


To derive the ELBO for \ipVAE, we replace the standard normal prior in \requation{eq:elbo-reconstruction-minus-priorKL} with the decoupled prior defined in \requation{eq:realNVP}. Using the change of variable formula, the KL divergence term in \requation{eq:elbo-reconstruction-minus-priorKL} can be simplified into
\footnote{Complete derivation in Appendix \ref{sec:KL-derivation}.}:


\begin{align}
    \KL{\qphi(\z|\x)}{p(\z)} = -\E{\qphi(\z|\x)}{\sum_{k=1}^K \sum_{l=1}^L \bkl{k}{l} s_k\left(\bkl{k}{l} \zkl{k}{l}\right)} \nonumber\\
    - \frac{1}{2} \log |\SigmaZX| + \E{\qphi(\z|\x)}{\geta{\z}^T\geta{\z}} \label{eq:ipKL}
\end{align}
\noindent where $L$ is the latent dimension, $K$ is number of invertible blocks used to define the decoupled prior (see \requation{eq:gz}), $s_k$ is the scaling network of the $k-$th block, $\SigmaZX$ is the covariance matrix of the variational posterior $\qphi(\z|\x)$ (which is typically assumed to be diagonal), and $\bkl{k}{l}$ and $\zkl{k}{l}$ are the $l-$th element in $\bk{k}$ and $\zk{k}$ vectors, respectively.


\subsection{\ipVAE~ in Concert with VAE Regularizers}

The KL divergence in \requation{eq:ipKL} can be directly used for any regularized ELBO.  
However, there are some regularized models such as $\beta$-TCVAE \cite{chen2018isolating}, and InfoVAE \cite{zhao2019infovae} that introduce additional terms other than $\KL{\qphi(\z|\x)}{p(\z)}$ with $p(\z)$. These regularizers need to be modified when used with decoupled priors\footnote{The ELBO definitions for these regularizers can be found in Appendix \ref{sec:elbodef}}. 

\vspace{0.05in}
\noindent \textbf{$\beta$-\ipVAE:} For $\beta$-VAE (both $\beta$-VAE-H \cite{higgins2017beta} and $\beta$-VAE-B \cite{burgess2018understanding} versions), the only difference in the ELBO \requation{eq:elbo-reconstruction-minus-priorKL} is reweighting the KL-term and the addition of certain constraints without 
introducing any additional terms. 
Hence, $\beta$-\ipVAE~ will retain the same reweighting and constraints, and only modify the KL divergence term according to \requation{eq:ipKL}.

\vspace{0.05in}
\noindent \textbf{Factor-\ipVAE:} FactorVAE \cite{kim2018disentangling} introduces a total correlation term  $\KL{\qphi(\z)}{\qphi(\bar{\z})}$ to the ELBO in \requation{eq:elbo-reconstruction-minus-priorKL}, where $\qphi(\bar{\z}) = \prod_{l=1}^L \qphi(\zl{l})$ and $\zl{l}$ is the $l-$th element of $\z$. This term promotes disentanglement of the latent dimensions of $\z$, impacting the representation learning aspect of VAE. Hence, in the case of the decoupled prior, the total correlation term should be applied to the \emph{representation space}. In this sense, the decoupled prior only affects the KL divergence term as described in \requation{eq:ipKL} for the Factor-\ipVAE~ model.

\vspace{0.05in}
\noindent  \textbf{$\beta$-TC-\ipVAE}: Regularization provided by $\beta$-TCVAE \cite{chen2018isolating} factorizes the ELBO into the individual latent dimensions based on the decomposition given in \cite{hoffman2016elbo}. The only term that includes $p(\z)$ is the KL divergence between marginals, \ie $\KL{\qphi(\z)}{p(\z)}$. This term in  $\beta$-TCVAE is assumed to be factorized and is evaluated via sampling, facilitating the direct incorporation of the decoupled prior.
%
In particular, we can just sample from the base distribution $\zo \sim p(\zo) = \stdNrmZo$ and compute the corresponding sample $\z \sim p(\z)$ using $\z = \ginveta{\zo}$.

\vspace{0.05in}
\noindent  \textbf{Info-\ipVAE}: In InfoVAE \cite{zhao2019infovae}, the additional term in the ELBO is again the divergence between aggregate posterior and the prior, \ie  $\KL{\qphi(\z)}{p(\z)}$. This KL divergence term is replaced by different divergence families; adversarial training \cite{makhzani2015adversarial}, Stein variational gradient \cite{liu2016stein}, and maximum-mean discrepancy MMD \cite{gretton2007kernel,li2015generative,dziugaite2015training}. However, adversarial-based divergences can have unstable training and Stein variational gradient scales poorly with high dimensions \cite{zhao2019infovae}. Motivated by the MMD-based results in \cite{zhao2019infovae}, we focus here on the MMD divergence to evaluate this marginal divergence.  
%
For Info-\ipVAE, we start with the ELBO of InfoVAE and modify the standard KL divergence term using \requation{eq:ipKL}. In addition, we compute the marginal KL divergence using MMD, which quantifies the divergence between two distributions by comparing their moments through sampling. Similar to $\beta$-TC-\ipVAE, we can sample from $p(\zo)$ and use the inverse mapping to compute samples in the $\z-$space.  


%% file: results_v2.tex
\section{Experiments}\label{sec:results}
\begin{figure}
    \centering
    \includegraphics[width=\linewidth]{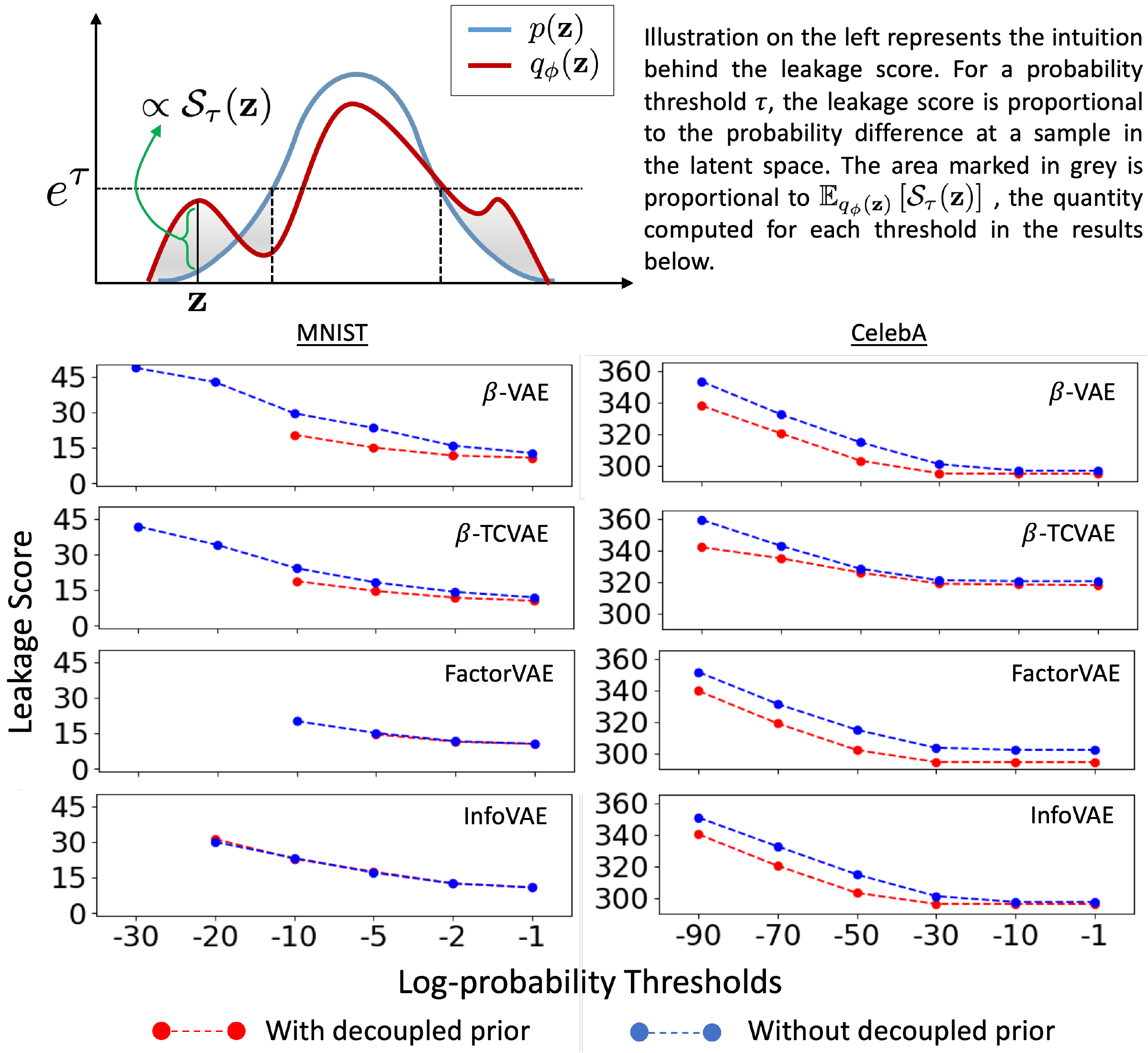}
    \caption{\textbf{\ipVAEs~ have less latent leaks:} Leakage scores for regularized VAEs on MNIST and CelebA data (missing values mean there are no samples with $\log(p(\z)) < \tau $, implying zero leakage at that threshold). }
    \label{fig:hipos_metric}
\end{figure}

\begin{table*}[!h]
\centering
\caption{Generative metrics (\textit{\textbf{lower} is better}) for vanilla VAE and regularized VAEs using standard normal and decoupled priors. \textbf{FID} = Fréchet Inception Distance. \textbf{LP} = Low-Posterior. \textbf{sKL} = symmetric KL divergence. \textbf{NLL} = Negative Log-Likelihood}
\resizebox{\textwidth}{!}{%
\begin{tabular}{|c||c|c|c|c||c|c|c|c||c|c|c|c|}
\hhline{-||----||----||----}
\multirow{2}{*}{\textbf{Methods}} & \multicolumn{4}{c||}{\textbf{MNIST}} & \multicolumn{4}{c||}{\textbf{SVHN}} & \multicolumn{4}{c|}{\textbf{CelebA}} \\ \hhline{|~||====||====||====|}

  & FID & FID (LP) & sKL & NLL ($\times 10^3$) & FID & FID (LP) & sKL & NLL ($\times 10^3$) & FID & FID (LP) & sKL & NLL ($\times 10^3$) \\ \hhline{=||====||====||====|}

VAE \cite{kingma2014auto,rezende2014stochastic} & 137.4  &  165.0   & 1.26   &  3.56  & 78.9 & 83.8 & 53.67 & 0.386 & 81.4 & 79.0 & 59.3 & 9.26   \\ \hhline{-||----||----||----}

\ipVAE          & \textbf{129.0}   &  \textbf{153.1}  & \textbf{0.88}   & \textbf{ 3.53}  & \textbf{50.8} & \textbf{55.2} & \textbf{13.02} & \textbf{0.318} & \textbf{71.5} & \textbf{74.3} & \textbf{10.4} & \textbf{4.91}     \\ \hhline{=||====||====||====|}

$\beta$-VAE-H \cite{higgins2017beta}  & 144.2   & 163.1   & 4.49   &  4.12  & 96.7 & 97.6 & 10.35 & 0.611 & 80.3 & 79.9 & 39.7 & \textbf{6.93}     \\ \hhline{-||----||----||----}

$\beta$-\ipVAE-H  & \textbf{127.1}  & \textbf{127.4}   & \textbf{1.07}   &  \textbf{2.98}  & \textbf{65.2} & \textbf{67.7} & \textbf{4.05} & \textbf{0.592} & \textbf{67.2} & \textbf{72.5} & \textbf{33.5} & 10.6 \\ \hhline{=||====||====||====|}

$\beta$-VAE-B \cite{burgess2018understanding}  & 130.8   & 163.6   & 2.74   & 3.11   & 61.7 & 68.5 & 2.62 & 0.606 & 75.7 & 79.6 & 25.8 & 12.5  \\ \hhline{-||----||----||----} 

$\beta$-\ipVAE-B  & \textbf{113.3} & \textbf{114.1}   & \textbf{1.32}   & \textbf{2.80}   & \textbf{51.1} & \textbf{50.3} & \textbf{2.47} & \textbf{0.550} & \textbf{67.9} & \textbf{72.0} & \textbf{19.1} & \textbf{10.4} \\ \hhline{=||====||====||====|}

$\beta$-TCVAE  \cite{chen2018isolating} & 149.8   & 200.3   & 4.48   & 2.91   & 69.2 & 70.5 & 7.76 & 9.86 & 83.8 & 83.0 & 93.6 & \textbf{9.33}   \\ \hhline{-||----||----||----}

$\beta$-TC-\ipVAE  & \textbf{133.3} & \textbf{133.1}  & \textbf{2.07}   &\textbf{ 2.70}   & \textbf{50.3} & \textbf{53.8} & \textbf{2.94} & \textbf{4.52} & \textbf{80.3} & \textbf{81.4} & \textbf{90.3} & 10.0 \\ \hhline{=||====||====||====|}

FactorVAE    \cite{kim2018disentangling}  & 130.5   &  191.2  & 1.04   &  \textbf{3.50}   & 97.2 & 108.5 & 1.91 & \textbf{2.13} & 82.6 & 86.8 & 71.3 & \textbf{9.89}  \\ \hhline{-||----||----||----}

Factor-\ipVAE   & \textbf{120.8}   & \textbf{121.3}   & \textbf{0.85}   & 3.60       & \textbf{86.3} & \textbf{86.9} & \textbf{1.57} & 2.36 & \textbf{65.0} & \textbf{73.4} & \textbf{51.3} & 12.2 \\ \hhline{=||====||====||====|}

InfoVAE    \cite{zhao2019infovae}     & 128.7    & 133.2  & 2.89  & 2.88   & 81.3 & 83.2 & 4.91 & \textbf{1.55} & 76.5 & 79.1 & 30.6 & \textbf{11.1}   \\ \hhline{-||----||----||----}

Info-\ipVAE     & \textbf{110.1}   & \textbf{110.5}   &\textbf{ 1.70}   & \textbf{2.81}   & \textbf{62.9} & \textbf{67.7} & \textbf{2.67} & 1.56 & \textbf{68.9} & \textbf{72.9} & \textbf{20.3} & 12.1  \\ \hhline{-||----||----||----}
\end{tabular}}
\label{tab:Gen-Metric}
\end{table*}

\begin{figure*}
    \centering
    \includegraphics[width=\textwidth]{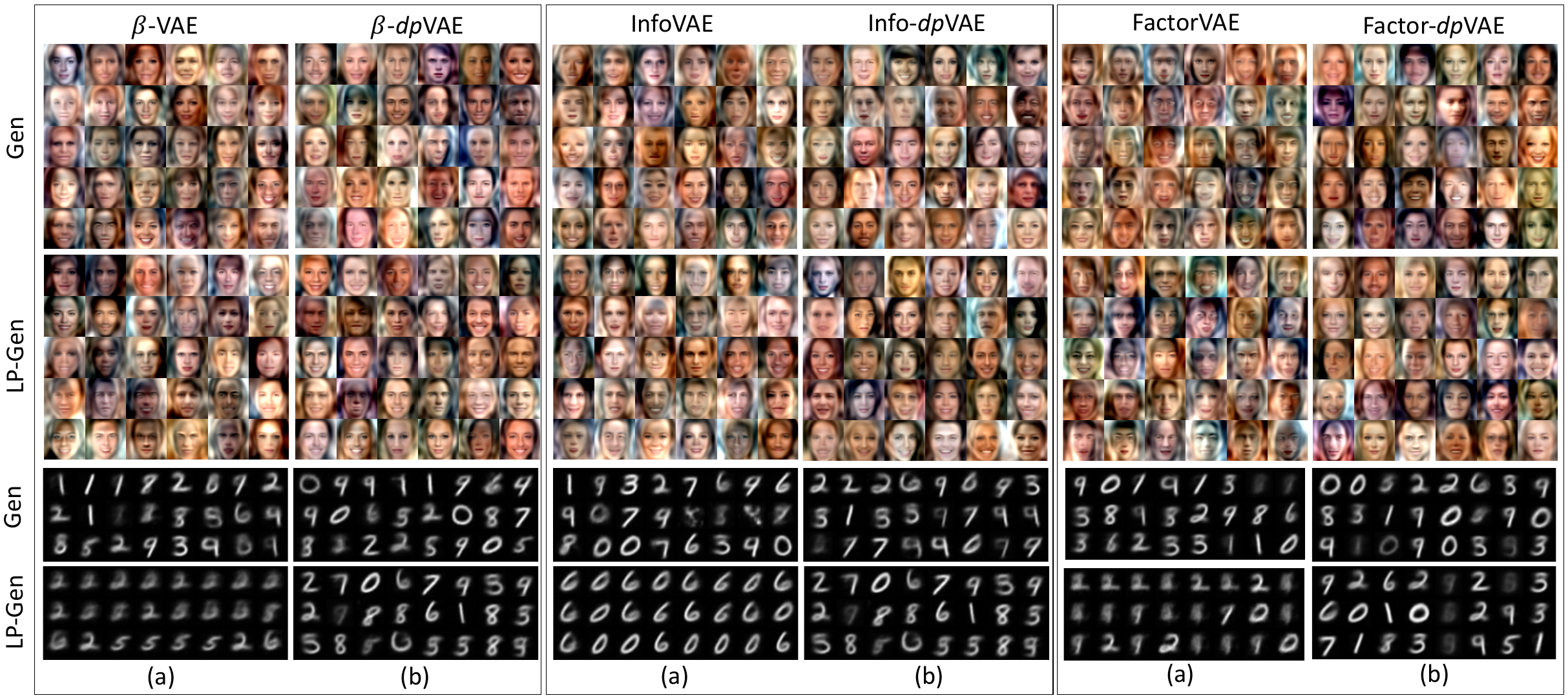}
    \caption{\textbf{\ipVAEs~ generate low-posterior samples with better quality:} We show generated (Gen) and low-posterior (LP-Gen) samples for CelebA and MNIST on regularized VAEs (a) without and (b) with decoupled priors.}
    \label{fig:genFigures}
\end{figure*}
We experiment with three benchmark image datasets, namely MNIST \cite{lecun2010mnist}, SVHN \cite{netzer2011reading}, CelebA (cropped version) \cite{liu2015deep}. 
We train these datasets with 
VAE \cite{kingma2014auto,rezende2014stochastic} and five regularized VAEs,
namely $\beta$-VAE-H \cite{higgins2017beta}, $\beta$-VAE-B \cite{burgess2018understanding}, $\beta$-TCVAE) \cite{chen2018isolating}, FactorVAE \cite{kim2018disentangling} and InfoVAE \cite{zhao2019infovae}. 
%
We showcase, qualitatively and quantitatively, that \ipVAEs~ improve sample generation while retaining the benefits of representation learning provided by the regularizers\footnote{Architecture, hyperparameter and train/test split details used for each dataset are described in Appendix \ref{sec:hyperparameters}.}.
%

\subsection{Generative Metrics}

%

We use the following quantitative metrics to assess the generative performance of the regularized VAEs with the decoupled prior in contrast to the standard normal prior.

\vspace{0.05in}
\noindent\textbf{Fréchet Inception Distance (FID):} The FID score is based on the statistics, assuming Gaussian distribution, computed in the feature space defined using the inception network features \cite{szegedy2016inception}. 
FID score quantifies both the sample diversity and quality. 
Lower FID means better sample generation. 

\vspace{0.05in}
\noindent\textbf{Symmetric KL Divergence (sKL):} To quantify the overlap between $p(\z)$ and $\qphi(\z)$ in the representation space ($p(\z)$ being the decoupled prior for \ipVAEs~ or the standard normal), we compute $\operatorname{sKL} \!\!\!= \KL{p(\z)}{\qphi(\z)} + \KL{\qphi(\z)}{p(\z)}$ through sampling (using 5,000 samples). sKL also captures the existence of pockets and leaks in $\qphi(\z)$. Lower sKL implies there is a better overlap between $p(\z)$ and $\qphi(\z)$, indicating better generative capabilities. 

\vspace{0.05in}
\noindent\textbf{Negative Log-likelihood:} We estimate the likelihood of held-out samples under a trained model using importance sampling (with 21,000 samples) proposed in \cite{rezende2014stochastic}, where $\operatorname{NLL} \!\!\!=\!\!\! \log \E{\qphi(\z|\x)}{\ptheta(\x|\z)p(\z)/\qphi(\z|\x)}$. Lower NLL means better sample generation since the learned model supports unseen samples drawn from the data distribution. 

\begin{figure}
    \centering
    \includegraphics[width=\linewidth]{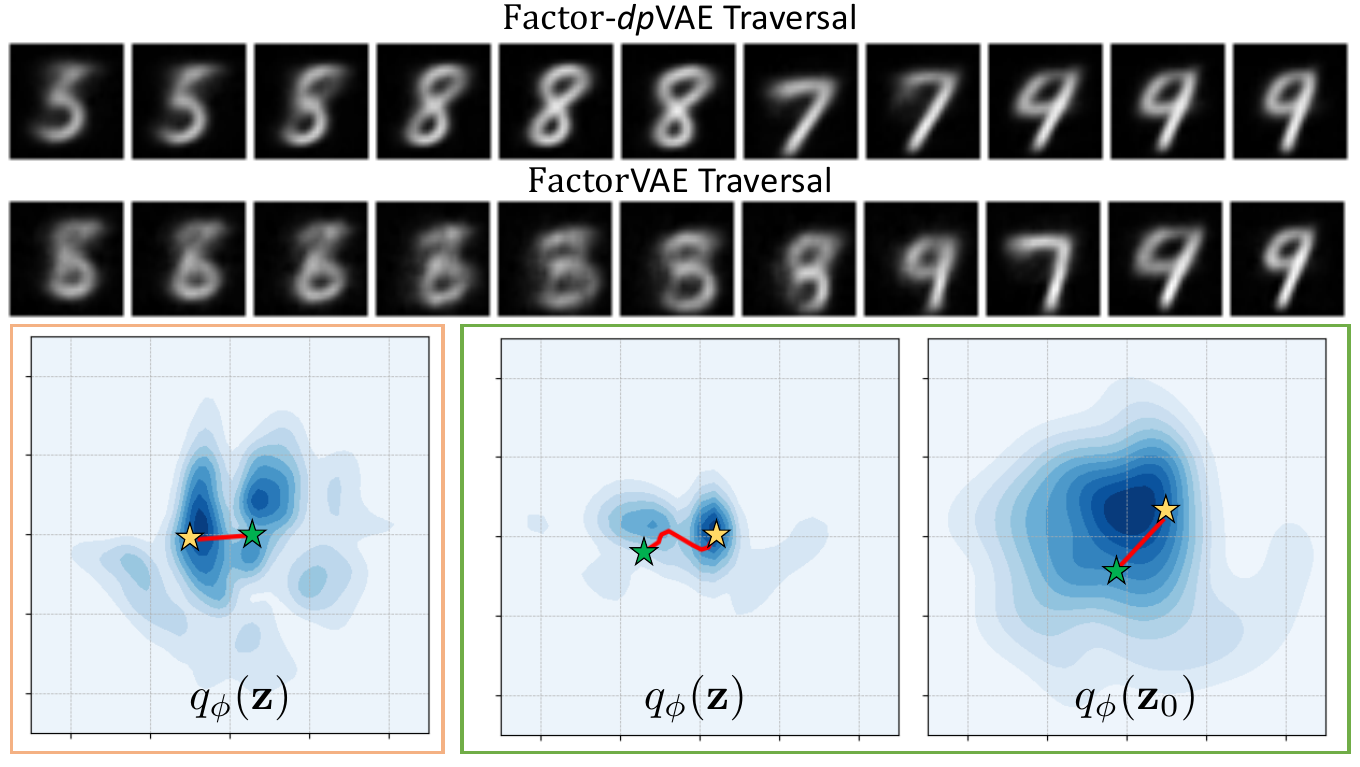}
    \caption{\textbf{Latent traversal for \ipVAEs~ does not path through latent pockets:} The top rows showcases latent traversal for FactorVAE and Factor-\ipVAE~ on MNIST data. The \textcolor{orange}{orange box} is the $\qphi(\z)$ for FactorVAE and the red line shows the traversal between starting and ending  points (green and yellow stars, respectively). The \textcolor{green}{green box} shows the \textit{same} traversal in $\qphi(\z_0)$ that is mapped using $g_\eta^{-1}$ to the representation space, demonstrated using $\qphi(\z)$. We see that the traversal path in $\qphi(\z)$ tries to avoid low probability regions, which correspond to better image quality.}
    \label{fig:mnist_traverse}
\end{figure}

\vspace{0.05in}
\noindent \textbf{Leakage Score:} To assess the effect of decoupled priors on latent leaks (as manifested by high posterior samples), we devise a new metric based on log-probability differences. We sample from the aggregate posterior $\z \sim \qphi(\z)$. If $\log(p(\z)) < \tau$, where $\tau \in \mathbb{R}$ is a chosen threshold value, then we consider the sample to lie in a ``leakage region" defined by $\tau$. This sample is considered a high-posterior sample at the $\tau-$level since the sample is better supported under the aggregate posterior than the prior (see illustration in \rfigure{fig:hipos_metric}).
%
Based on the threshold value, these leakage regions are less likely to be sampled from. In order to not lose significant regions from the data manifold, we want the aggregate posterior corresponding to these samples to attain low values as well.
%
%
%
To quantify latent leakage for a trained model, we propose a \emph{leakage score} as $\operatorname{LS}(\tau) = \mathbb{E}_{q_\phi(\mathbf{z})} \left[\mathcal{S}_\tau(\mathbf{z}) \right]$, where for a given $\z \sim \qphi(\z)$ at a particular threshold $\tau$, $\mathcal{S}_{\tau}(\z)$ is defined as:
\begin{align}
    \mathcal{S}_{\tau}(\z) = \begin{cases} 
      \log\left(\frac{\qphi(\z)}{p(h(\z))}\right) & \log(p(\z)) < \tau \ \\
      0 & \log(p(\z)) \geq \tau
   \end{cases}
\end{align}
Here, $h$ is the identity function in VAEs with standard normal prior, and is $g_\eta$ for \ipVAEs. 

\subsection{Generation Results and Analysis}

In Table \ref{tab:Gen-Metric}, we observe that \ipVAEs~ perform better than their corresponding regularized VAEs without the decoupled prior for each dataset. When comparing VAEs with and without decoupled priors (\eg InfoVAE and Info-\ipVAE), we use the same hyperparameters and perform no additional tuning. This showcases the robustness of the decoupled prior \wrt hyperparameters, facilitating its direct use with any regularized VAE. 
We report the FID scores on both the randomly generated samples from the prior and the low-posterior samples. As analyzed in section \ref{sec:submanifold}, if the low-posterior samples lie on the data manifold, then the learned latent space is pocket-free. 
Results in Table \ref{tab:Gen-Metric} suggest that for all \ipVAEs, the FID scores for the randomly generated samples and low-posterior ones are comparable, suggesting that all the pockets in the latent space are filled. 
Qualitative results of sample generation for CelebA and MNIST  are shown in \rfigure{fig:genFigures}. We show both the random prior and  low-posterior sample generation with and without the decoupled prior for three different regularizations. The quality of generated samples using \ipVAEs~ is better or on par with those without the decoupled prior. But more importantly, one can observe a significant quality improvement in the low-posterior samples, which aligns with the quantitative results in Table \ref{tab:Gen-Metric}. 
%
In \rfigure{fig:hipos_metric}, we report the leakage score $\operatorname{LS}(\tau)$
as a function of log-probability thresholds for different regularizers with and without the decoupled priors. We observe that \ipVAEs~ result in models with lower latent leakage. 
This is especially true at lower thresholds, which suggests that even when $p(\z)$ is small, the $\qphi(\z)$ is small as well, preventing the loss of significant regions from the data manifold. 



\subsection{Latent Traversals Results}

We perform latent traversals between samples in the latent space. 
We expect that in VAEs, there will be instances of the traversal path crossing the latent pockets resulting in poor intermediate samples. In contrast, we expect \ipVAEs~ will map the linear traversal in $\z_0$ (generation space) to a non-linear traversal in $\z$ (representation space), while  avoiding low probability regions. 
This is observed for MNIST data traversal ($L=2$) and is depicted in \rfigure{fig:mnist_traverse}. 

\begin{figure}
    \centering
    \includegraphics[width=\linewidth]{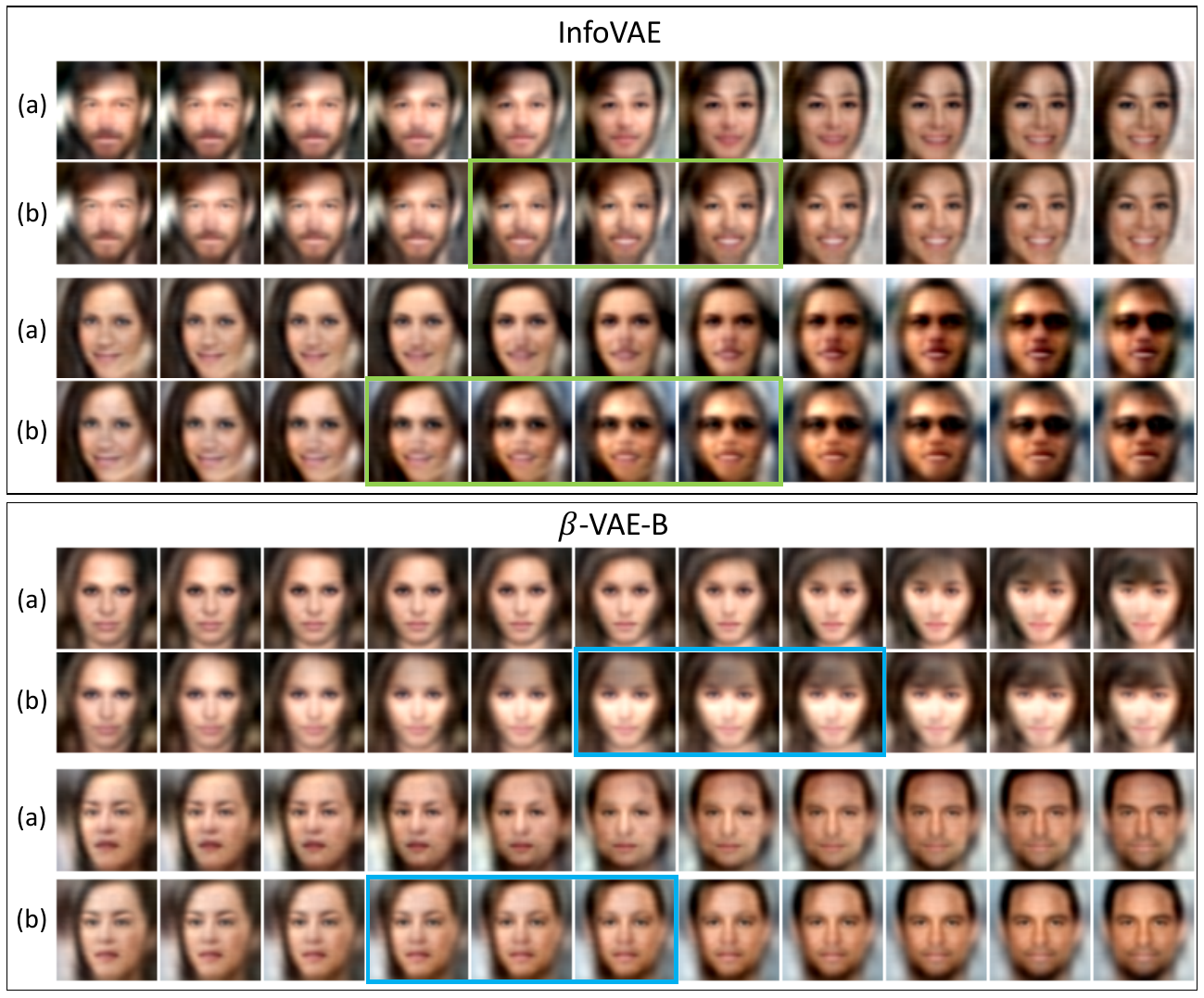}
    \caption{\textbf{Latent traversals on CelebA:} (a) with and (b) without decoupled prior. We notice for InfoVAE, the \textcolor{green}{green boxes} highlight the faces with unrealistic deformations, such as excessive teeth in the second row and noisy images in the fourth row.  In comparison, Info-\ipVAE~ results in a more smooth traversal. We see similar effects from $\beta$-VAE, with odd shadowing due to hair and face rotation, highlighted in \textcolor{blue}{blue boxes}. Again, $\beta$-\ipVAE~ makes these transitions smooth and seamless.}
    \label{fig:abtraverse}
\end{figure}

\begin{figure}
    \centering
    \includegraphics[width=\linewidth]{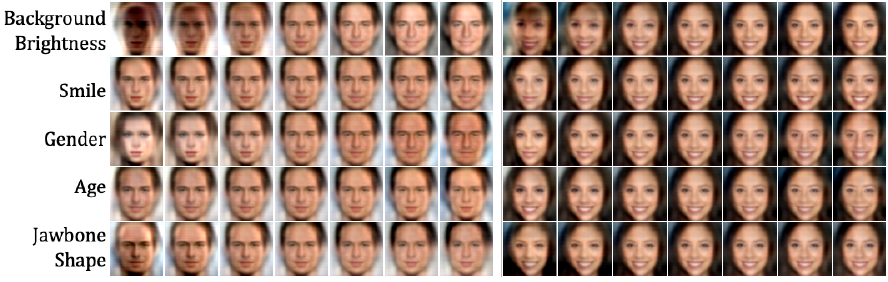}
    \caption{\textbf{Factor-\ipVAE~ latent traversals across the top five latent dimensions sorted by their standard deviation from $\qphi(\z)$:} 
    Traversals start with the reconstructed image of a given sample and move $\pm 5$ standard deviations along a latent dimension. Results from other \ipVAEs~ similarly retain the latent space disentanglement.
    %
}
    \label{fig:factortraverse}
\end{figure}

We also qualitatively observe similar occurrences in CelabA traversals (see \rfigure{fig:abtraverse}). 
Finally, we want to attest that the addition of the decoupled prior to a regularizer does not affect it's ability to improve the latent representation.  We demonstrate this by observing latent factor traversals 
for CelebA trained on Factor-\ipVAE, where we vary one dimension of the latent space while fixing the others.  One can observe that Factor-\ipVAE~ is able to isolate different attributes of variation in the data, as shown in \rfigure{fig:factortraverse}.

%% file: conclusion.tex
\section{Conclusion}\label{sec:conclusion}
In this paper, we define and analyze the submanifold problem for regularized VAEs, or the tendency of a regularizer to accentuate the creation of pockets and leaks in the latent space. This submanifold structure manifests the mismatch between the aggregate posterior and the latent prior which in turn causes degradation in generation quality. 
To overcome this trade-off between sample generation and latent representation, we propose the decoupled prior family as a general regularization framework for VAE and demonstrate its efficacy on \sota VAE regularizers. 
We demonstrate that \ipVAEs~ generate better quality samples as compared with their standard normal prior based counterparts, via qualitative and quantitative results. 
Additionally, we qualitatively observe that the representation learning (as improved by the regularizer) is not adversely affected by \ipVAEs.
Decoupled priors can act as a pathway to realizing the true potential of VAEs as both a representation learning and a generative modeling framework.
%
Further work in this direction will include exploring more expressive inference and generative models (\eg PixelCNN \cite{van2016conditional}) in conjuction with decoupled priors.
We also believe more sophisticated invertible architectures (\eg. RAD \cite{dihn2019RAD}) and base distributions will provide further improvements.


%% file: appendix.tex
\textit{This appendix includes the derivation of \ipVAE~ training objective, the ELBO definitions of \sota VAE regularizers with the decoupled prior, and experimental details (architecture, hyperparameters, and train/test splits) for MNIST, SVHN and CelebA experiments. }

\section{Derivation of $\KL{\qphi(\z|\x)}{p(\z)}$ with Decoupled Prior}
\label{sec:KL-derivation}

Consider a bijective mapping between $\mcZ$ and $\mcZo$ defined by a function $\geta{\z} = \zo$. The \textit{change of variable formula} for mapping probability distribution on $\z$ to $\zo$ is given as follows:
\begin{align}\label{eqn:changeformula}
p(\z) = p(\zo)\bigg |\frac{\partial \zo}{\partial \z}\bigg | = p(\geta{\z})\bigg | \frac{\partial \geta{\z}}{\partial \z}\bigg |
\end{align}
The $g-$bijection is parameterized by $K$ \textit{affine coupling layers}, each layer is a bijection block $\zk{k-1} = \getakz{k}{\zk{k}}$ of the form,
\begin{eqnarray}
\getakz{k}{\zk{k}}&=& \bk{k} \odot \zk{k}  \nonumber \\ 
&+& (1 - \bk{k}) \odot \left[ \zk{k} \odot \exp \left(s_k(\bk{k} \odot \zk{k}) \right) \right. \nonumber \\ 
&& ~~~~~~~~~~~~~~~~~~~ \left. + t_k\left(\bk{k} \odot \zk{k}\right)\right], \label{eq:invblock}
\end{eqnarray}
\noindent where $\z = \zk{K}$, $\odot$ is the Hadamard (\ie element-wise) product,
$\bk{k} \in \{0,1\}^L$ is a binary mask used for partitioning the $k-$th block input, and $\eta = \{s_1, ..., s_K, t_1, ..., t_K\}$ are the deep networks parameters of the scaling $s_k$ and translation $t_k$ functions of the $K$ blocks. 
Stacking these affine coupling layers constitutes the functional mapping between the representation and the generation spaces. The $g-$bijection is thus defined as,
\begin{align}
    \geta{\z} = \zo = \getak{1} \circ \dots \circ \getak{K-1} \circ \getakz{K}{\z} . \label{eq:gz}
\end{align}
The KL divergence is given as,
\begin{eqnarray}
\operatorname{KL} &:=& \KL{\qphi(\z|\x)}{p(\z)}   \nonumber \\
&=& \int\qphi(\z|\x) \left[\log(\qphi(\z|\x)) - \log(p(\z))\right]d\z .
\end{eqnarray}
Using the change of variable formula in \requation{eqn:changeformula}, we have the following.
\begin{eqnarray}
\operatorname{KL} &=& \int\qphi(\z|\x)\bigg[\log(\qphi(\z|\x)) - \log(p(\geta{\z})) \nonumber\\ 
&& ~~~~~~~~~~~~~ - \log\left(\bigg | \frac{\partial \geta{\z}}{\partial \z}\bigg |\right)\bigg]d\z \\
 &=& \KL{\qphi(\z|\x)}{p(\geta{\z})} \nonumber \\ 
 && ~~~~~~~~~~~~~ - \E{\qphi(\z|\x)}{\log\left(\bigg | \frac{\partial \geta{\z}}{\partial \z}\bigg |\right)} .
\label{eq:full}
\end{eqnarray}

\noindent The first term in \requation{eq:full} can be derived as follows,
 \begin{align}
    T_1 &= \KL{\qphi(\z|\x)}{p(\geta{\z})} \\ 
    &= \int \qphi(\z|\x)\log\left(\frac{\qphi(\z|\x)}{p(\geta{\z})}\right) d\z \\ 
    &= -\mathbb{H}(\qphi(\z|\x)) - \int \qphi(\z|\x)\log\left(p(\geta{\z}\right) d\z,
\end{align}
\noindent where $\mathbb{H}(\qphi(\z|\x))$ is the differential entropy of the variational posterior distribution. This approximate posterior is a multivariate Gaussian with a diagonal covariance matrix, \ie $\qphi(\z|\x) \sim \NrmZ{\muZX}{\SigmaZX}$, where $\SigmaZX = \sigmaZXdiag$, and $\sigmaZX \in \RLp$. 
This results in a closed-form expression for the entropy of the approximate posterior \cite{matrixcookbook}, given as: 
\begin{align}
    \mathbb{H}(\qphi(\z|\x)) = \frac{L}{2} + \frac{L}{2}\log(2\pi) + \frac{\log|\SigmaZX|}{2}. \label{eq:entropy}
\end{align}
The probability of the base latent space (\ie the generation space) in the decoupled prior is assumed to be a standard normal distribution, \ie $p(\geta{\z}) = \mathcal{N}(\geta{\z}; 0, \mathbb{I}_L)$. Together with \requation{eq:entropy} and ignoring constants terms, the first term in \requation{eq:full} can be simplified into,
\begin{eqnarray}
    T_1 &=&  -\frac{\log|\SigmaZX|}{2} + \int \qphi(\z|\x)\frac{L}{2}\log\left(\frac{1}{2\pi}\right) d\z  \nonumber\\ 
    &+& \frac{1}{2}\int \qphi(\z|\x) \geta{\z}^T \geta{\z}d\z \\ 
    &=& \frac{1}{2}\bigg[-\log{|\SigmaZX|} + const \nonumber \\
    &+&  \E{q(\z|\x)}{\geta{\z}^T\geta{\z}}\bigg] 
    \label{eq:term-1}
\end{eqnarray}

\noindent The second term in \requation{eq:full} can be derived as follows,
 \begin{eqnarray}
    T_2 &=& \E{\qphi(\z|\x)}{\log{\left(\bigg |\frac{\partial \geta{\z}}{\partial \z}\bigg |\right)}}  \nonumber \\
    &=& \int \qphi(\z|\x) \log{\left(\bigg |\frac{\partial \geta{\z}}{\partial \z}\bigg |\right)}d\z.
\end{eqnarray}

\noindent Applying the chain rule to $\geta{\z}$, which is a composition of functions as defined in \requation{eq:gz}, and combining this with the multiplicative property of determinants, we have the following,
\begin{align}
\bigg |\frac{\partial \geta{\z}}{\partial \z}\bigg | = \prod\limits_{k = 1}^K \bigg |\frac{\partial \getakz{k}{\zk{k}}}{\partial \zk{k}}\bigg |.
\end{align}
Hence, we have the second term in \requation{eq:full} can be expressed as follows:
\begin{align}
T_2 &= \E{\qphi(\z|\x)}{\sum\limits_{k = 1}^K\log{\left(\bigg |\frac{\partial \getakz{k}{\zk{k}}}{\partial \zk{k}}\bigg |\right)}}.
\end{align}
Similar to \cite{dinh2017density}, deriving the Jacobian of individual affine coupling layers yields to an upper triangular matrix. Hence, the determinant is simply the product of the diagonal values, resulting in,
\begin{align}
\log\left(\bigg |\frac{\partial \getakz{k}{\zk{k}}}{\partial \zk{k}}\bigg |\right) = \sum\limits_{l = 1}^Lb^l_ks_k(b^l_kz^l_k). \label{eq:logdet}
\end{align}
Using this simplification, we finally have an expression for the second term expressed as,
\begin{align}
T_2 = \E{\qphi(\z|\x)}{\sum\limits_{k=1}^K\sum\limits_{l = 1}^Lb^l_ks_k(b^l_kz^l_k)}.
\label{eq:term-2}
\end{align}

Substituting \requation{eq:term-2} and \requation{eq:term-1} in \requation{eq:full}, the KL divergence term of the decoupled prior can be written as,
\begin{eqnarray}
\operatorname{KL} &=& \frac{1}{2}\bigg[-\log{|\SigmaZX|} +  \E{q(\z|\x)}{\geta{\z}^T\geta{\z}}\bigg] \nonumber \\
&-& \E{\qphi(\z|\x)}{\sum\limits_{k=1}^K\sum\limits_{l = 1}^Lb^l_ks_k(b^l_kz^l_k)}
\end{eqnarray}

\section{ELBOs for Different Regularizers}
\label{sec:elbodef}
In this section, we define the ELBO (\ie training objective) for individual regularizers considered (see section 4.3 in the main manuscript) and how they differ under the application of decoupled prior. Here, we only serve to provide more mathematical clarity of these modifications.

\vspace{0.1in}
\noindent \textbf{$\beta$-\ipVAE}: The ELBO for the  $\beta$-VAE \cite{higgins2017beta} can be expressed as follows:
\begin{align}
    \mcL(\theta,\phi) &= \En{p(\x)}{ \E{\qphi(\z|\x)}{\log \ptheta(\x|\z)}  \nonumber \\  &- \beta \KL{\qphi(\z|\x)}{p(\z)}}
\end{align}
By substituting the KL divergence under the decoupled prior, the ELBO for $\beta$-\ipVAE~ can defined as follows:
\begin{align}
    \mcL(\theta,\phi, \eta) &= \En{p(\x)}{ \E{\qphi(\z|\x)}{\log \ptheta(\x|\z)}  \nonumber \\   &- \frac{\beta}{2}\big[-\log{|\SigmaZX|} + \E{q(\z|\x)}{\geta{\z}^T\geta{\z}}\big] \nonumber \\ &- \beta\E{\qphi(\z|\x)}{\sum\limits_{k=1}^K\sum\limits_{l = 1}^Lb^l_ks_k(b^l_kz^l_k)}}
\end{align}

The alternate formulation, $\beta$-VAE-B \cite{burgess2018understanding}, has a similar ELBO function with some additional parameters applied to the KL divergence term. 

\vspace{0.1in}
\noindent\textbf{Factor-\ipVAE}: 
The ELBO for FactorVAE \cite{kim2018disentangling} is given as follows:
\begin{align}
    \mcL(\theta,\phi) &= \En{p(\x)}{ \E{\qphi(\z|\x)}{\log \ptheta(\x|\z)}  \nonumber \\  &- \KL{\qphi(\z|\x)}{p(\z)}} - \gamma\KL{\qphi(\z)}{\qphi(\bar{\z})}
\end{align}
To modifying this ELBO with the decoupled prior, the KL divergence term between the posterior and prior that contains $p(\z)$ is the only term that needs to be reformulated. This results in:
\begin{align}
    \mcL(\theta,\phi, \eta) &= \En{p(\x)}{ \E{\qphi(\z|\x)}{\log \ptheta(\x|\z)}  \nonumber \\   &- \frac{1}{2}[-\log{(|\SigmaZX|)} + \E{q(\z|\x)}{\geta{\z}^T\geta{\z}}} \nonumber \\ &- \E{\qphi(\z|\x)}{\sum\limits_{k=1}^K\sum\limits_{l = 1}^Lb^l_ks_k(b^l_kz^l_k)}] \\ &- \gamma\KL{\qphi(\z)}{\bar{\qphi(\z)}}
\end{align}

\vspace{0.1in}
\noindent \textbf{$\beta$-TC-\ipVAE:} 
Following similar notation as provided in \cite{chen2018isolating}, the ELBO for $\beta$-TCVAE \cite{chen2018isolating} is given as follows:
\begin{align}
    \mcL(\theta,\phi) &= \E{p(n)}{ \E{\qphi(\z|n)}{\log \ptheta(n|\z)}}  \nonumber \\  &- \alpha I_{\qphi}(\z; n) - \beta \KL{\qphi(\z)}{\prod\limits_{j}\qphi(z_j)} \nonumber \\ &- \gamma \sum\limits_{j}\KL{\qphi(z_j)}{p(z_j)}
\end{align}
Due to the factorized representation of the prior, the last KL divergence is computed via sampling. Hence, the modification for the decoupled prior becomes trivial. We simply sample from $\z_0$ space (which is assumed to be factorized) and pass it through $\getainv{-1}(z_{0j})$ to obtain a sample in $\z$ space, the ELBO can thus be modified as follows:
\begin{align}
    \mcL(\theta,\phi, \eta) &= \E{p(n)}{ \E{\qphi(\z|n)}{\log \ptheta(n|\z)}}  \nonumber \\  &- \alpha I_{\qphi}(\z; n) - \beta \KL{\qphi(\z)}{\prod\limits_{j}\qphi(z_j)} \nonumber \\ &- \gamma \sum\limits_{j}\KL{\qphi(z_j)}{p(\getainv{-1}(z_{0j}))}
\end{align}

\vspace{0.1in}
\noindent \textbf{Info-\ipVAE}: 
For InfoVAE \cite{zhao2019infovae}, the ELBO (using the MMD divergence) is given as follows: 
\begin{align}
    \mcL(\theta,\phi) &= \En{p(\x)}{ \E{\qphi(\z|\x)}{\log \ptheta(\x|\z)}  \nonumber \\  &- (1 - \alpha)\KL{\qphi(\z|\x)}{p(\z)}} 
    \nonumber \\
    &- (\alpha + \lambda - 1) D_{MMD}(\qphi(\z) || p(\z))
\end{align}
The modification for the decoupled prior takes place in two terms; the MMD divergence term, which is computed via sampling from the aggregate posterior, and the prior in $\z$ space, which acts on the representation space in the decoupled prior. Therefore, the modification is very similar to the one applied in $\beta$-TCVAE. Additionally, the KL divergence term will be modified normally. The final ELBO is thus as follows:

\begin{align}
    \mcL(\theta,\phi, \eta) &= \En{p(\x)}{ \E{\qphi(\z|\x)}{\log \ptheta(\x|\z)}  \nonumber \\   &- \frac{1 - \alpha}{2}[-\log{(|\SigmaZX|)} + \E{q(\z|\x)}{\geta{\z}^T\geta{\z}}] \nonumber \\ &- (1 - \alpha) \E{\qphi(\z|\x)}{\sum\limits_{k=1}^K\sum\limits_{l = 1}^Lb^l_ks_k(b^l_kz^l_k)}} \nonumber \\
    &- D_{MMD}(\qphi(\z)||p(\getainv{-1}(\zo)))
\end{align}

\section{Architectures and Hyperparameters}
\label{sec:hyperparameters}
In this section, we give more details for the architecture and hyperparameters used and the data handling for the four different datasets used in the paper, namely two moons toy data, MNIST, SVHN, and CelebA. These are also described for different regularizers used on each of these datasets.
\begin{figure}
    \centering
    \includegraphics[width=\linewidth]{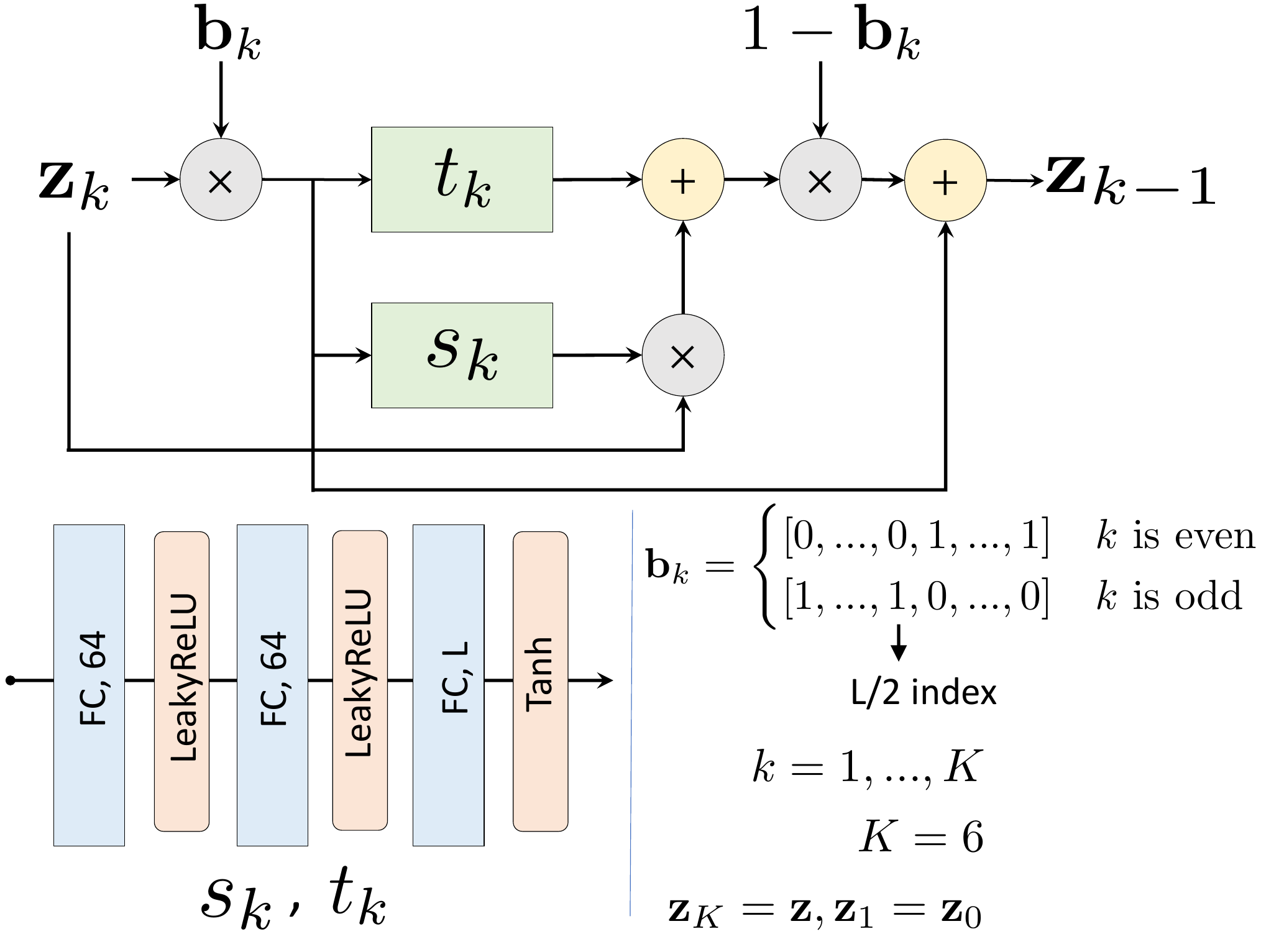}
    \caption{ \textbf{Architecture details of the decoupled prior.} Architecture description of individual affine coupling blocks and the details of binary masks. The top shows the structure of an individual affine coupling layer representing the function $\getakz{k}{\zk{k}}$. Sndividual scaling $s_k$ and translation $t_k$ functions have the same architecture for all the block (bottom-left). Bottom-right shows the binary masks and number of affine coupling layers.}
    \label{fig:affineCoup}
\end{figure}

\begin{figure*}
\includegraphics[width=\linewidth]{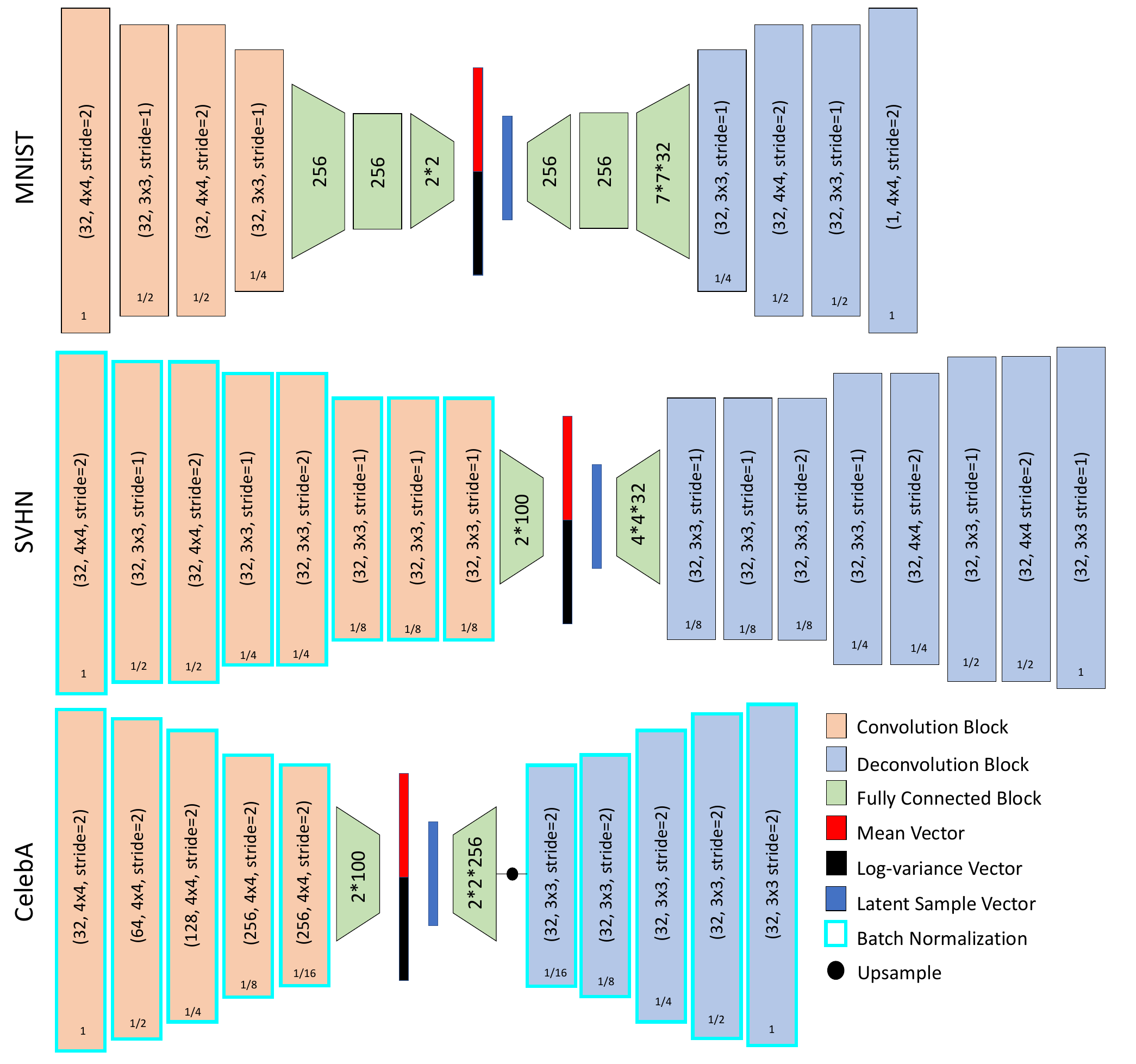}
\caption{ \textbf{Architecture description for different datasets.} This figure shows the VAE architecture for MNIST, SVHN, and CelebA datasets. This architecture is kept the same for all the regularizations with and without the decoupled prior.
}
\label{fig:otherarchs}
\end{figure*}

Figure \ref{fig:otherarchs} illustrates the VAE architecture for all the datasets and the regularizers reported in the experiments section of the paper. For MNIST and SVHN, we use ReLU as the non-linear activation function, and for CelebA we use leaky ReLU \cite{maas2013rectifier}. For the two moons data, the VAE architecture consists of two fully connected layers of size 100 and 50 (from input to latent space) in the encoder. The decoder is a mirrored version of the encoder.
We use two-dimensional latent space for the two moons data. 
The architecture that is added for the decoupled prior is the same for all the experiments we present in the paper.  This architecture for the affine coupling layers that connects $\z$ and $\zo$ is shown in \rfigure{fig:affineCoup}.

FactorVAE \cite{kim2018disentangling} has a discriminator architecture, which has five fully connected layers each with 1000 hidden units. Each fully connected layer is followed by a leaky ReLU activation of negative slope of 0.2. This discriminator architecture is the same for all experiments, except for the changing input size (\ie the latent dimension $L$).

The learning rate for all the experiments was set to be $10^{-4}$, and the batch size for MNIST was 100, SVHN and CelebA were 50. We execute all the experiments for 100,000 iterations (100,000/B epochs where B is the batch size). No other pre-processing was performed while conducting these experiments. The regularization specific hyperparameters are mentioned in Table \ref{tab:regparam}. These hyperparameters were kept the same for all datasets.

\begin{table}[]
\centering
\caption{\textbf{Table representing hyperparameters for individual regularizations}. These hyperparameters were set to be the same with and without the decoupled prior.}
\label{tab:regparam}
\begin{tabular}{|l|l|}
\hline
\textbf{Methods} & \textbf{Parameters} \\ \hline
$\beta$-VAE-H  \cite{higgins2017beta}    &  $\beta = 4$                    \\ \hline
$\beta$-VAE-B  \cite{burgess2018understanding}      & $\gamma = 15$, $C_{max} = 25$, $C_{stop} = 100000$                     \\ \hline
$\beta$-TC-VAE  \cite{chen2018isolating}      & $\alpha = 1$, $\beta = 4$, $\gamma = 15$                    \\ \hline
FactorVAE  \cite{kim2018disentangling}      &   $\gamma = 1000$                  \\ \hline
InfoVAE     \cite{zhao2019infovae}     &  $\alpha = 0$, $\lambda = 1000$                   \\ \hline
\end{tabular}
\end{table}